\documentclass[letterpaper]{article} % DO NOT CHANGE THIS
\usepackage{aaai23}  % DO NOT CHANGE THIS
\usepackage{dsfont}
\usepackage{times}  % DO NOT CHANGE THIS
\usepackage{helvet}  % DO NOT CHANGE THIS
\usepackage{courier}  % DO NOT CHANGE THIS
\usepackage[hyphens]{url}  % DO NOT CHANGE THIS
\usepackage{graphicx} % DO NOT CHANGE THIS
\usepackage{natbib}  % DO NOT CHANGE THIS AND DO NOT ADD ANY OPTIONS TO IT
\usepackage{caption} % DO NOT CHANGE THIS AND DO NOT ADD ANY OPTIONS TO IT
\usepackage{algorithm}
\usepackage{algorithmic}
\usepackage{newfloat}
\usepackage{listings}
\DeclareCaptionStyle{ruled}{labelfont=normalfont,labelsep=colon,strut=off} % DO NOT CHANGE THIS
\floatstyle{ruled}
\newfloat{listing}{tb}{lst}{}
\floatname{listing}{Listing}
\title{In-game Toxic Language Detection: Shared Task and Attention Residuals}
\author{
    Yuanzhe Jia, Weixuan Wu, Feiqi Cao, Soyeon Caren Han
}
\affiliations{
    School of Computer Science, University of Sydney, Australia\\
    \{yjia5612, wewu8797, fcao0492\}@uni.sydney.edu.au, caren.han@sydney.edu.au
}

\usepackage{bibentry}
\begin{document}

\maketitle

\begin{abstract}
In-game toxic language becomes the hot potato in the gaming industry and community. There have been several online game toxicity analysis frameworks and models proposed. However, it is still challenging to detect toxicity due to the nature of in-game chat, which has extremely short length. In this paper, we describe how the in-game toxic language shared task has been established using the real-world in-game chat data. In addition, we propose and introduce the model/framework for toxic language token tagging (slot filling) from the in-game chat. The code is publicly available on GitHub\footnote{\url{https://github.com/Yuanzhe-Jia/In-Game-Toxic-Detection}}.
\end{abstract}

\section{Introduction}
Toxic behaviour has become a severe problem in recent online games and the gaming industry. The nature of the toxic utterances in the in-game chat is very different from the other online domain, such as social media or online news. It has a much shorter length because game players tend to type in-game chat during playing. The longer utterances occur only in pre- or post-game discussions. With this in-game chat nature in mind, understanding the slot(word token)-level~\cite{8} is crucial to detect the in-game toxic language. In this paper, we describe the established slot(token)-based in-game toxic language detection shared task, and present the best model and its novel components. The result shows the novelty of the best model by comparing the baselines from the CONDA~\cite{1}.

\section{Shared Task and Dataset}
\label{sec:sharedtask}

We set up a shared task competition\footnote{\url{https://www.kaggle.com/competitions/2022-comp5046-a2}} for performing a sequence token labelling task to identify the type of semantics conveyed by each slot in-game chat utterances provided by CONDA~\cite{1}. CONDA consists of 44,869 utterances from chat logs of 1,921 Dota2 matches, with 26,921/8,974/8,974 utterances for training/validation/test sets respectively. CONDA provides 6 distinct slot labels: \textbf{T} (Toxicity), \textbf{C} (Character), \textbf{D} (Dota-specific), \textbf{S} (Game Slang), \textbf{P} (Pronoun) and \textbf{O} (Other) to provide a deeper understanding of game context. A total of 312 student teams participated in our shared task. Each team was allowed 20 submissions per day. In total, we received 3,646 submissions across 4 weeks.
Most participants pre-processed the provided data simply by tokenizing the utterances into slots by spaces, as CONDA provides the cleaned utterances to provide slot labels. 
There are also teams applying data augmentation to increase the number of instances they can have for training.
%\subsection{Input Embedding}\label{sec:input_embedding}
For the input embedding, combinations of syntactic, semantics and domain-related embeddings were included in the participants' trials. For the syntactic embedding, most teams chose to encode POS tags and/or dependency parsing results based on models provided by SpaCy. For the semantic embedding, participants would either directly use pre-trained GloVE/FastText embedding vectors, or train their own FastText/Word2Vec based on the provided game chat corpus. Some teams tried to find a Dota-related corpus or toxicity detection dataset collected from social media to train the word embedding and inject more domain-related information into the model. A variety of sequence labelling model architectures are explored by the participants, and we will introduce the best-performing team's method and results.

\section{Methodology}
\label{sec:methodology}
Our model achieved the best performance from the shared task. It includes Bi-LSTM cells, attention residuals, label forcing and CRF. Since the model uses global information extracted from the attention mechanism as residuals to supplement bi-directional features, it is named Bi-directional Representations with Attention Residuals (BRAR).

\begin{figure}[t]
\centering
\includegraphics[width=0.6\columnwidth]{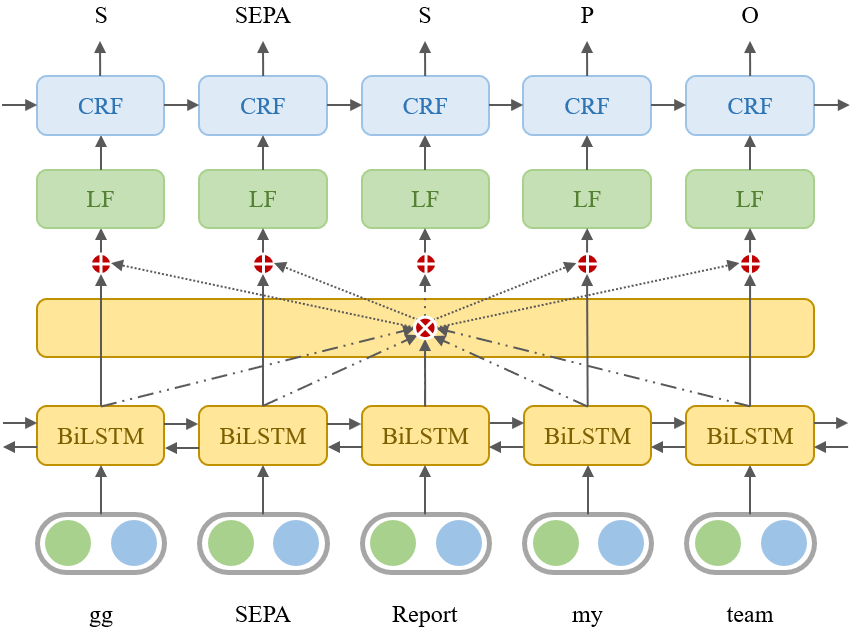}
\caption{The overall architecture of BRAR. The dotted arrows represent the proposed Attention Residuals, and the green boxes(LF) present the label-forcing components.}
\label{fig:structure}
\end{figure}

%\subsection{BiLSTM}
Bi-LSTM cells perform feature extraction on the input sequence $ x=(x_1,x_2,…,x_t) $ where $ t $ denotes the number of tokens. $ y=(y_1,y_2,…,y_t) $ is the corresponding slot labels with $ c $ unique values.
The attention layer aims to understand the global information of the input utterance. Attention 
$ a \in \mathds{R}^{2\lambda} $ 
is calculated in the following equation where
$ h_t \in \mathds{R}^{2\lambda} $ 
is the last hidden state of Bi-LSTM cells,
$ W_a \in \mathds{R}^{{2\lambda} \times {t}} $ 
is a trainable weight matrix and 
$ H \in \mathds{R}^{{t} \times {2\lambda}} $ 
is the output of Bi-LSTM:
\begin{equation}\label{eq2}
a = softmax(h_t \cdot W_a \cdot H) 
\end{equation}

As the global information is not always helpful for understanding each input token, it is treated as the residual of the token-level representation to form the feature 
$ f_i \in \mathds{R}^{{t} \times {c}} $,
while $ \alpha $ is a trainable parameter to scaling and
$ W_f \in \mathds{R}^{{2\lambda} \times {c}} $ 
is a trainable weight matrix for dimension transformation. When the global information benefits the token-level interpretation, it will speed up the convergence; otherwise, it will not reduce the prediction performance.
\begin{equation}\label{eq3}
f_i = W_f(H + \alpha \times a_i)
\end{equation}

%\subsection{Label Forcing (LF)}
The feature $ f_i $ is enhanced at the label forcing layer to form the emission score, which will be passed to the CRF layer to predict the tag of each token. For example, “gg” stands for “good game”. If all words “gg” in the training set are classified as “S” (Game Slang),  the word in the test set will have a high probability of being classified as “S”. In this case, if the probability information of the correspondence between words and labels in the training set can be learned, it will be of great help in predicting the results of the test set. Therefore, the proposed model creatively uses a novel technique: label forcing (LF). Specifically, the label distribution probability of each token in the training corpus is calculated and normalized by the following equation, where $ C_{ij} $ is the frequency of that the token $ i $ is annotated as the label $ j $. The overall architecture can be found in the Figure~\ref{fig:structure}. 
\begin{equation}\label{eq4}
p_i = C_{ij} / \sum_j C_{ij}
\end{equation}

\section{Experiment}
\label{sec:experiment}

The proposed model adopts FastText-50 as the input embeddings. The Bi-LSTM layer contains a hidden size of 5 in each direction, and the layer number is set to 1. The SGD optimizer has been applied with the learning rate of 0.1 and the weight decay of 1e-4. In addition, the training epoch is set to 2 with a batch size of 1 (same as the CONDA). As for evaluation metrics, the overall micro F1 and that for each slot label have been reported on the test set, but the O tag is excluded when calculating the overall F1-Score. The 5 baselines are selected based on CONDA paper. From Table \ref{tab:comparison} we can conclude that BRAR outperforms other baselines on most F1, especially in predicting T, S, and D labels.

\begin{table}
\centering
\scalebox{0.68}
{
\begin{tabular}{cccc cccc}
\hline\noalign{\smallskip}	
Model & F1 & F1(T) & F1(P) & F1(S) & F1(D) & F1(C) & F1(O) \\
\noalign{\smallskip}\hline\noalign{\smallskip}
RNN-NLU ~\shortcite{3} & 97.0 & 93.1 & 98.1 & 93.0 & 71.8 & 99.1 & 98.7 \\
Slot-gated~\shortcite{4} & 99.1 & 97.8 & 99.2 & 98.2 & 95.2 & 99.7 & 99.4 \\
Inter-BiLSTM~\shortcite{5} & 86.5 & 87.1 & 88.9 & 86.9 & 78.8 & 94.2 & 92.4 \\
Capsule NN~\shortcite{6} & 99.1 & 97.5 & 99.1 & 98.2 & 94.9 & 99.7 & 99.4 \\
Joint BERT (2019) & 98.9 & 97.2 & 99.2 & 97.9 & 91.4 & \textbf{99.8} & 99.3 \\
BRAR (our model) & \textbf{99.9} & \textbf{98.6} & \textbf{99.4} & \textbf{99.4} & \textbf{98.1} & 99.0 & \textbf{99.5} \\
\noalign{\smallskip}\hline
\end{tabular}
}
\caption{Overall performance with baselines (\%).}
\label{tab:comparison} 
\end{table}

We evaluated the model with different number of Bi-LSTM stacks/layers. The results shown in Table \ref{tab:ablation3} indicate that more layers lead to the lower F1 scores, though there is no significant difference among the three different number of layers. Therefore, we propose to use only 1 Bi-LSTM layer as it is computationally more efficient with a smaller number of parameters.

\begin{table}
\centering
\scalebox{0.68}
{
\begin{tabular}{cccc cccc}
\hline\noalign{\smallskip}	
Number of Stacks & F1 & F1(T) & F1(P) & F1(S) & F1(D) & F1(C) & F1(O) \\
\noalign{\smallskip}\hline\noalign{\smallskip}
1 layer & \textbf{99.9} & \textbf{98.6} & \textbf{99.4} & \textbf{99.4} & \textbf{98.1} & \textbf{99.0} & \textbf{99.5} \\
2 layers & 99.6 & 98.1 & 99.2 & 99.3 & 96.4 & 98.1 & \textbf{99.5} \\
3 layers & 98.6 & 96.5 & 99.1 & 98.3 & 81.8 & 94.4 & 99.4 \\
\noalign{\smallskip}\hline
\end{tabular}
}
\caption{Ablation for different stack layers (\%).}
\label{tab:ablation3} 
\end{table}

\section{Conclusion}
\label{sec:conclusion}

In this paper, we established a shared task for slot (word token)-based in-game toxic language detection, as well as the winner (the best model) in the shared task, which integrates bi-directional representations, attention residuals, the label forcing technique and CRF. Experiments indicate that the proposed best model is more effective in capturing global information between the semantic components in the slot filling task than existing models.

\section{Acknowledgments}
This work was supported by FortifyEdge. We would like to specially thank Mr Hyunsuk (David) Chung for his help and guidance for this research work.

\bibliography{aaai23}

\begin{thebibliography}{6}
\providecommand{\natexlab}[1]{#1}

\bibitem[{Bing and Ian(2016)}]{3}
Bing, L.; and Ian, L. 2016.
\newblock Attention-based recurrent neural network models for joint intent detection and slot filling.
\newblock In \emph{Interspeech 2016}, 685–689.

\bibitem[{Chenwei et~al.(2019)Chenwei, Yaliang, Nan, Wei, and S}]{6}
Chenwei, Z.; Yaliang, L.; Nan, D.; Wei, F.; and S, P., Yu. 2019.
\newblock Joint slot filling and intent detection via capsule neural networks.
\newblock In \emph{Proceedings of the 57th Annual Meeting of the ACL}, 5259– 5267.

\bibitem[{Chih-Wen et~al.(2018)Chih-Wen, Guang, Yun-Kai, Chih-Li, Tsung-Chieh, Keng-Wei, and Yun-Nung}]{4}
Chih-Wen, G.; Guang, G.; Yun-Kai, H.; Chih-Li, H.; Tsung-Chieh, C.; Keng-Wei, H.; and Yun-Nung, C. 2018.
\newblock Slot-gated modeling for joint slot filling and intent prediction.
\newblock In \emph{NAACL-HLT 2018}, 753– 757.

\bibitem[{Weld et~al.(2021{\natexlab{a}})Weld, Huang, Lee, Zhang, Wang, Guo, Long, Poon, and Han}]{1}
Weld, H.; Huang, G.; Lee, J.; Zhang, T.; Wang, K.; Guo, X.; Long, S.; Poon, J.; and Han, C. 2021{\natexlab{a}}.
\newblock CONDA: a CONtextual Dual-Annotated dataset for in-game toxicity understanding and detection.
\newblock In \emph{Findings of the Association for Computational Linguistics: ACL 2021}, 2406--2416.

\bibitem[{Weld et~al.(2021{\natexlab{b}})Weld, Huang, Long, Poon, and Han}]{8}
Weld, H.; Huang, X.; Long, S.; Poon, J.; and Han, S.~C. 2021{\natexlab{b}}.
\newblock A survey of joint intent detection and slot filling models in natural language understanding.
\newblock \emph{ACM Computing Surveys (CSUR)}.

\bibitem[{Yu, Yilin, and Hongxia(2018)}]{5}
Yu, W.; Yilin, S.; and Hongxia, J. 2018.
\newblock A bi-model based rnn semantic frame parsing model for intent detection and slot filling.
\newblock In \emph{NAACL-HLT 2018}.

\end{thebibliography}

\end{document}